\documentclass[runningheads]{llncs}

\usepackage{eccv}

\usepackage{eccvabbrv}

\usepackage{graphicx}
\usepackage{booktabs}

\usepackage[accsupp]{axessibility}  % Improves PDF readability for those with disabilities.

\usepackage{color}
\usepackage{amssymb}
\usepackage{graphicx}
\usepackage{amsmath}
\usepackage{float}
\usepackage{diagbox}

\usepackage{hyperref}
\hypersetup{
    colorlinks=true, 
    urlcolor=eccvblue,
    pdfstartview=Fit, 
    breaklinks=true
}

\usepackage{orcidlink}

\begin{document}

\title{DIFFender: Diffusion-Based Adversarial Defense against Patch Attacks} 

\titlerunning{DIFFender}

\author{Caixin Kang\inst{1}\orcidlink{0009-0001-1924-9311} \and
Yinpeng Dong\inst{2,5} \and
Zhengyi Wang\inst{2,6} \and
Shouwei Ruan\inst{1}\and
Yubo Chen\inst{1} \and
Hang Su\inst{2,4\star}\and
Xingxing Wei\inst{1,3}\thanks{Corresponding authors.} }

\authorrunning{C.~Kang et al.}

% \institute{Institute of Artificial Intelligence, Beihang University, Beijing 100191, China \and
% Dept. of Comp. Sci. \& Tech., Institute for AI, BNRist Center, THBI Lab,
% Tsinghua-Bosch Joint ML Center, Tsinghua University, Beijing, 100084, China \and
% Hangzhou Innovation Institute, Beihang University, Hangzhou 311228, China  \and Zhongguancun Laboratory, Beijing, 100080, China \and
% RealAI \and ShengShu
% % \email{lncs@springer.com}\\
% % \url{http://www.springer.com/gp/computer-science/lncs} \and
% % ABC Institute, Rupert-Karls-University Heidelberg, Heidelberg, Germany\\
% % \email{\{abc,lncs\}@uni-heidelberg.de}
% }

\institute{
\mbox{\inst{1} Institute of Artificial Intelligence, Beihang University, Beijing 100191, China}
\mbox{\inst{2} Dept. of Comp. Sci. \& Tech., Institute for AI, BNRist Center, THBI Lab,}
\mbox{Tsinghua-Bosch Joint ML Center, Tsinghua University, Beijing, 100084, China}
\mbox{\inst{3} Hangzhou Innovation Institute, Beihang University, Hangzhou 311228, China}
\mbox{\inst{4} Zhongguancun Laboratory, Beijing, 100080, China \quad
\inst{5} RealAI \quad
\inst{6} ShengShu
}
\email{\{caixinkang,xxwei\}@buaa.edu.cn, \{dongyinpeng,suhangss\}@tsinghua.edu.cn}\\
% \footnotesize{\texttt{\{dongyinpeng,suhangss,dcszj\}@tsinghua.edu.cn, caixinkang@buaa.edu.cn, 1711302013@st.gxu.edu.cn}}
}

\maketitle

\begin{abstract}
Adversarial attacks, particularly patch attacks, pose significant threats to the robustness and reliability of deep learning models. Developing reliable defenses against patch attacks is crucial for real-world applications. This paper introduces DIFFender, a novel defense framework that harnesses the capabilities of a text-guided diffusion model to combat patch attacks. Central to our approach is the discovery of the Adversarial Anomaly Perception (AAP) phenomenon, which empowers the diffusion model to detect and localize adversarial patches through the analysis of distributional discrepancies. DIFFender integrates dual tasks of patch localization and restoration within a single diffusion model framework, utilizing their close interaction to enhance defense efficacy. Moreover, DIFFender utilizes vision-language pre-training coupled with an efficient few-shot prompt-tuning algorithm, which streamlines the adaptation of the pre-trained diffusion model to defense tasks, thus eliminating the need for extensive retraining. Our comprehensive evaluation spans image classification and face recognition tasks, extending to real-world scenarios, where DIFFender shows good robustness against adversarial attacks. The versatility and generalizability of DIFFender are evident across a variety of settings, classifiers, and attack methodologies, marking an advancement in adversarial patch defense strategies. Our
code is available at \url{https://github.com/kkkcx/DIFFender}.
\end{abstract}

% \vspace{-0.8cm}
\section{Introduction}
\label{sec:intro}
Deep neural networks are vulnerable to adversarial examples \cite{szegedy2013intriguing,goodfellow2014explaining}, in which imperceptible perturbations are intentionally added to natural examples, leading to incorrect predictions with high confidence of the model \cite{ma2021poisoning,xu2022rethinking}.
Most adversarial attacks and defenses are devoted to studying the $\ell_p$-norm threat models \cite{goodfellow2014explaining,carlini2017towards,dong2018boosting,madry2017towards}, which assume that the adversarial perturbations are restricted by the $\ell_p$ norm to be imperceptible. However, the classic $\ell_p$ perturbations require modification of every pixel of the images, which is typically not practical in the physical world. On the other hand, adversarial patch attacks \cite{brown2017adversarial,karmon2018lavan,li2021generative,wei2022adversarial}, which usually apply perturbations to a localized region of the objects, are more physically realizable. Adversarial patch attacks pose significant threats to real-world applications, such as face recognition \cite{Sharif2016Accessorize,xiao2021improving}, autonomous driving~\cite{jing2021too,zhu2023understanding,dong2023benchmarking,kong2024robodrive}.

Although many adversarial defenses against patch attacks have been proposed in the past years, the defense performance is not satisfactory, which cannot meet the demands of the safety and reliability of real-world applications.
Some methods employ adversarial training \cite{wu2019defending, rao2020adversarial, wei2024revisiting} and certified defenses \cite{gowal2019scalable, chiang2020certified}, which are only effective against specific attacks but generalize poorly to other forms of patch attacks in the real world \cite{nie2022diffusion}. Another category of patch defense is based on pre-processing techniques \cite{hayes2018visible,naseer2019local,yu2021defending,liu2022segment}, which usually destroy the patterns of adversarial patches by image completion or smoothing. However, these methods can hardly restore the images with high fidelity, leading to visual artifacts of the reconstructed images that impact recognition. They can also be evaded by stronger adaptive attacks due to gradient obfuscation \cite{athalye2018obfuscated}.

\begin{figure*}[t]
  \centering
  % \fbox{\rule[-.5cm]{0cm}{4cm} \rule[-.5cm]{4cm}{0cm}}
  \includegraphics[width=0.95\linewidth]{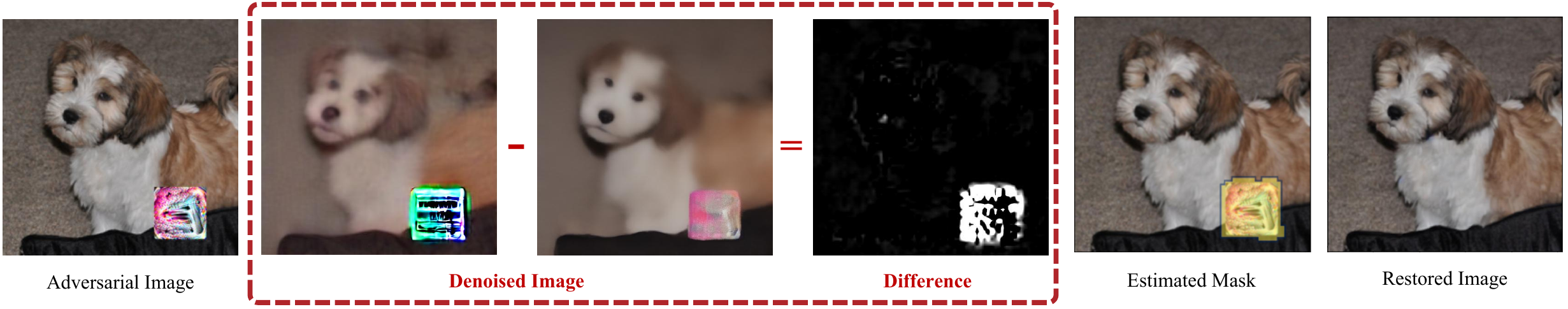}
  \caption{\small The intriguing phenomenon of the diffusion model. A diffusion model is performed multiple times on the given adversarial image, and the differences between any two denoised images are pronounced within the adversarial patch regions, which can be leveraged to further pinpoint the location of adversarial patches.}
 
  \label{fig:1}
\end{figure*}

Recently, diffusion models \cite{sohl2015deep,ho2020denoising} have emerged as a powerful family of generative models, and have been successfully applied to improving adversarial robustness by purifying the input data~\cite{nie2022diffusion,wang2022guided,xiao2022densepure}. 
Our initial intuition is to explore whether diffusion purification can defend against patch attacks. However, we find it fails to counter such attacks since it cannot remove the adversarial patches completely. In contrast, we discover the \textbf{Adversarial Anomaly Perception (AAP)} phenomenon, as shown in Fig.~\ref{fig:1}. The phenomenon suggests that we can calculate the difference between various denoised images to identify the region of adversarial patches. Subsequently, it facilitates targeted restoration of specific patch-affected areas. The reason behind the phenomenon may be that adversarial patches are often complexly crafted perturbations or contextually misplaced elements, significantly differing from the natural image distributions it is trained on.
This phenomenon indicates progress in understanding how diffusion models can differently respond to adversarial patches and resolves the inherent trade-off between purifying the adversarial patches and preserving the image semantics.

\begin{figure*}[tp]
  \centering
  % \fbox{\rule[-.5cm]{0cm}{4cm} \rule[-.5cm]{4cm}{0cm}}
  \includegraphics[width=0.74\linewidth]{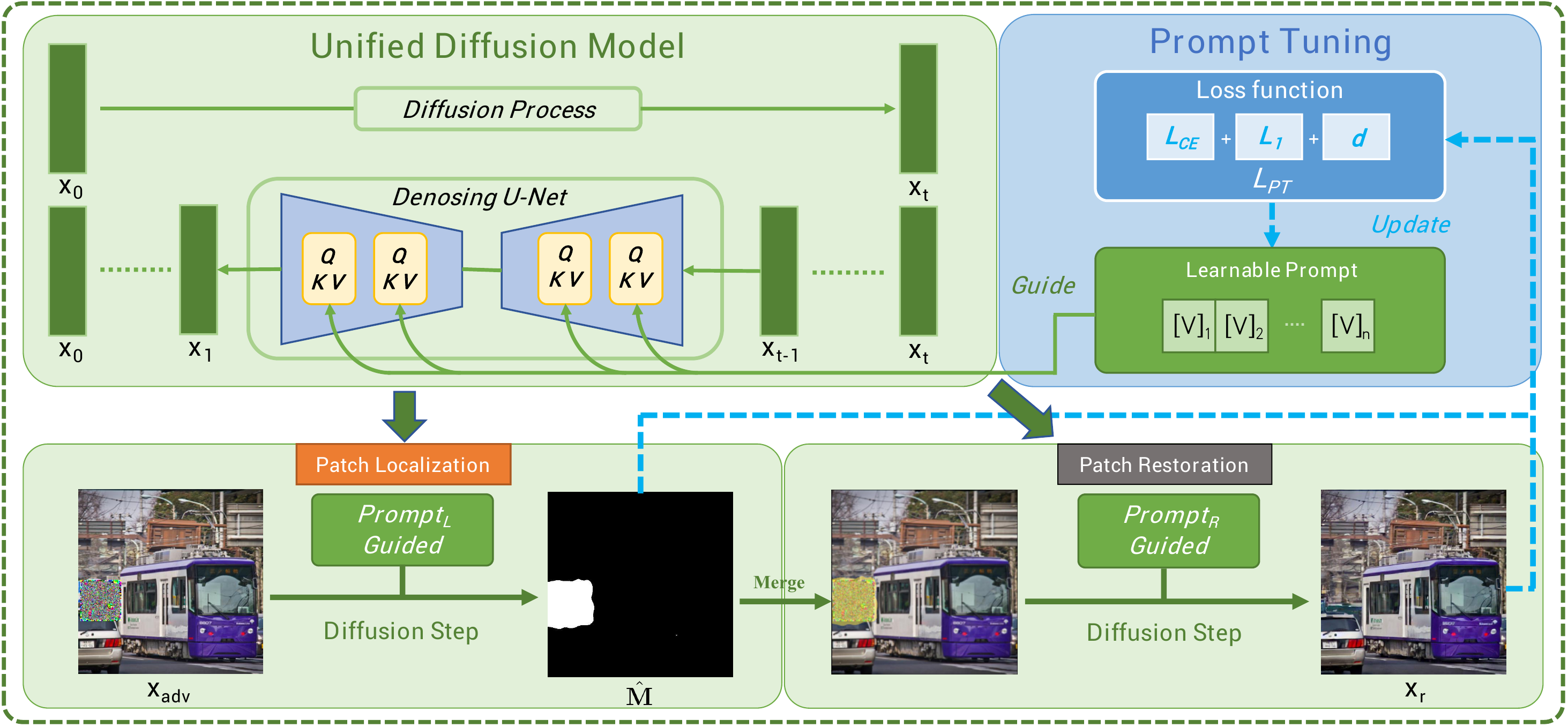}
  \caption{\small Pipeline of DIFFender. DIFFender leverages a unified diffusion model to jointly guide the localization and restoration of adversarial patch attacks, and combines a prompt-tuning module to facilitate efficient tuning.}
  \label{fig:2}
  % \vspace{-0.7cm}
\end{figure*}

Based on the AAP phenomenon, we propose \textbf{DIFFender}, a novel defense framework against adversarial patch attacks with pre-trained diffusion models. DIFFender localizes the region of the adversarial patch by comparing the differences between various denoised images and then recovers the identified patch region in the image while preserving the integrity of the underlying content. Importantly, these two stages are carefully guided by a unified diffusion model, thus we can utilize the close interaction between them to improve the whole defense (i.e., an accurate localization will promote the following restoration, and a perfect restoration will help evaluate the performance of localization step in return). Specifically, we incorporate a text-guided diffusion model such that DIFFender can localize and recover the adversarial patches more accurately with textual prompts. Moreover, we design a few-shot prompt-tuning algorithm to facilitate simple and efficient tuning, enabling the pre-trained diffusion model to easily adapt to the adversarial defense task for improved robustness.
The pipeline of DIFFender is illustrated in Fig.~\ref{fig:2}. In summary, our contributions are as follows:

\begin{itemize}

\item  We uncover the intriguing Adversarial Anomaly Perception (AAP) phenomenon within the diffusion model, enabling it to leverage the distributional discrepancies between adversarial patches and natural images for accurate localization, thus overcoming the trade-off between purifying patches and preserving image semantics. This approach broadens the applicability of diffusion techniques, making it feasible to employ the diffusion model in countering adversarial patch attacks.

\item Arising from the AAP phenomenon, we introduce DIFFender, an innovative diffusion-based defense framework. DIFFender employs one diffusion model throughout the entire process to both localize and restore adversarial patches, leveraging vision-language pre-training to implement efficient defense. To our knowledge, DIFFender stands as the first framework to leverage the diffusion model for comprehensive defense against patch attacks, marking a notable advancement in the field.

\item Additionally, we develop an efficient prompt-tuning module and design three effective losses. The losses fine-tune the pre-trained diffusion model through the tuning process, enabling the model to co-optimize both localization and restoration modules, thereby achieving improved defense performance. This approach not only enhances the model's adaptability but also reduces the computational overhead associated with traditional retraining methods.

\item  We conduct extensive experiments on image classification, face recognition, and further in the physical world, demonstrating that DIFFender effectively reduces the attack 
 success rate even under strong adaptive attacks. The results indicate that DIFFender can also generalize well to various scenarios, diverse classifiers, and multiple attack methods.

\end{itemize}

\section{Related work} 

\noindent\textbf{Adversarial attacks.} 
Deep neural networks (DNNs) can be misled to produce erroneous outputs \cite{szegedy2013intriguing,goodfellow2014explaining,dong2022viewfool} by introducing perturbations to input examples. Most adversarial attacks \cite{goodfellow2014explaining,moosavi2016deepfool,madry2017towards,dong2018boosting,ma2022tale} typically induce misclassification by adding small perturbations to the pixels of input examples. However, while these methods can effectively generate adversarial examples in the digital world, they lack practicality in the real world. 
On the other hand, adversarial patch attacks~\cite{brown2017adversarial,karmon2018lavan,li2021generative,wei2022adversarial,zhong2022shadows} aim to deceive models by applying a pattern or a sticker to a localized region of the object, which are more realizable in the physical world.

\noindent\textbf{Adversarial defenses.}
As attacks evolve, various defense methods have emerged. However, most existing defenses primarily focus on global perturbations with $\ell_p$ norm constraints, including former diffusion-based defenses~\cite{nie2022diffusion,wang2022guided,xiao2022densepure}, and defenses against patch attacks have not been extensively studied. 
Despite the effectiveness of adversarial training \cite{wu2019defending,rao2020adversarial,zhao2024mitigating} and certified defenses \cite{gowal2019scalable,chiang2020certified} against specific attacks, they have limited generalization to other patch attacks.

Therefore, most studies focus on pre-processing defenses. Digital Watermarking \cite{hayes2018visible} utilizes saliency maps to detect adversarial regions and employs erosion operations to remove small holes. Local Gradient Smoothing  \cite{naseer2019local} performs gradient smoothing on regions with high gradient amplitudes, taking into account the high-frequency noise introduced by patch attacks. Feature Normalization and Clipping \cite{yu2021defending} involves gradient clipping operations on images to reduce informative class evidence based on knowledge of the network structure. SAC~\cite{liu2022segment} defends against patch attacks by detecting and removing patches. Jedi \cite{tarchoun2023jedi} utilizes entropy to obtain masks for patches. However, these methods can hardly reconstruct the original image and can be evaded by adaptive attacks \cite{athalye2018obfuscated}. In contrast, we introduce the AAP phenomenon and propose to leverage pre-trained diffusion models to localize and restore the adversarial patches. This method not only addresses the limitations of existing defenses but also opens new pathways for research in using diffusion models for patch defense.

% \vspace{-0.4cm}
\section{Methodology}

In this section, we first introduce the discovery of the AAP phenomenon within diffusion models. Following this understanding, we outline the whole framework of DIFFender and detail the improved techniques by prompt tuning.

\begin{figure*}[h]
  \centering
  % \fbox{\rule[-.5cm]{0cm}{4cm} \rule[-.5cm]{4cm}{0cm}}
  \includegraphics[width=0.78\linewidth]{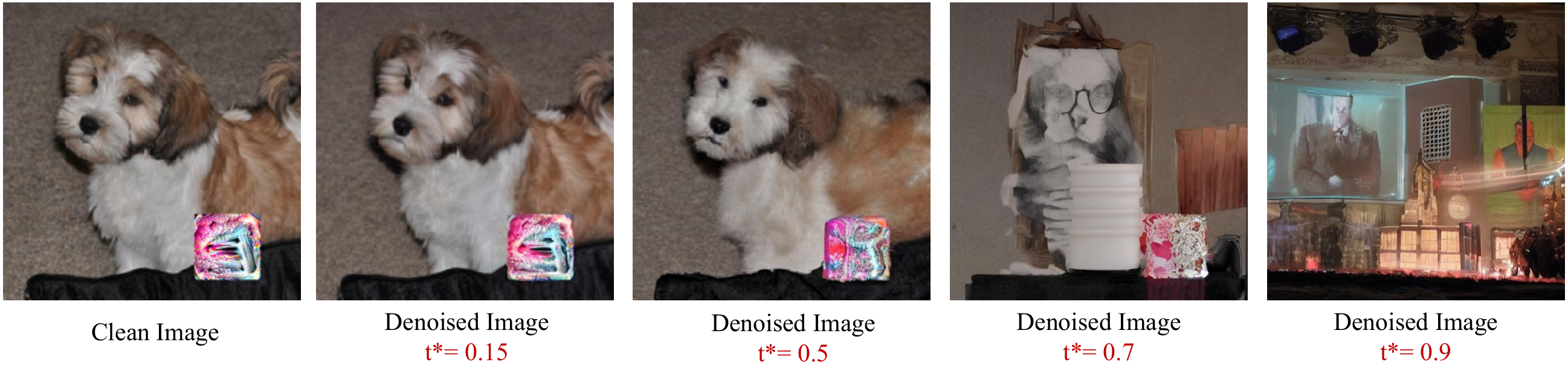}  
  \caption{\small Denoised results by diffusion model at different noise ratios. With small ratios ( \( t^* = 0.15/0.5 \) ), the patch cannot be purified; conversely, the global structure becomes lost with large ratios ( \( t^* = 0.7/0.9 \) ).}
  \label{fig:new}
\end{figure*}

\begin{figure*}[b]
  \centering
  \includegraphics[width=0.90\linewidth]{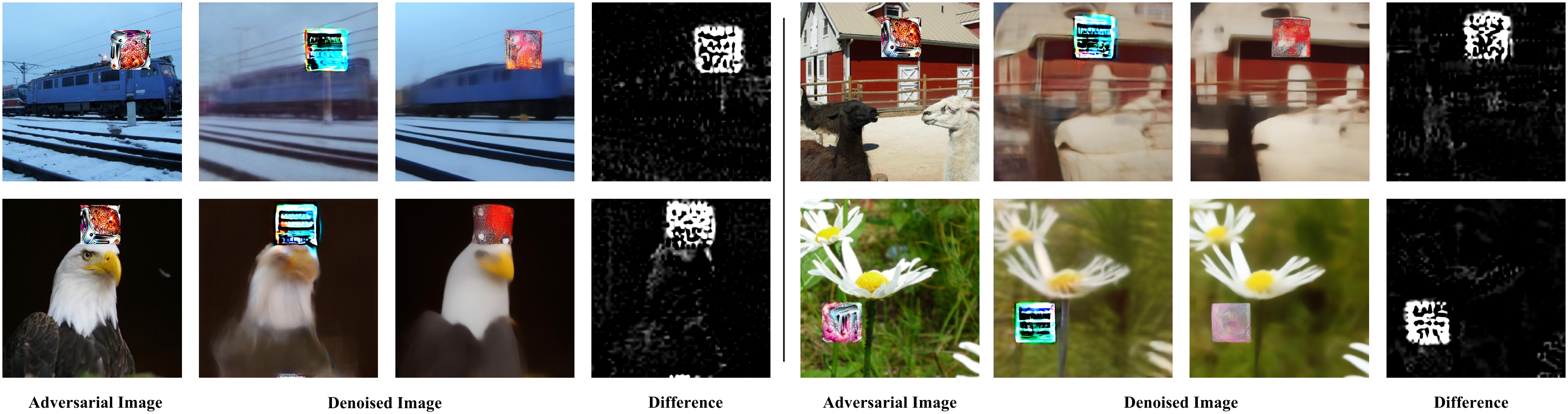}  
  \caption{\small In the analysis of ImageNet images, we find a pronounced difference specifically within regions affected by adversarial patches. This observation provides empirical evidence supporting the AAP phenomenon.}
  \label{fig:observation}
\end{figure*}

\subsection{Discovery of the AAP Phenomenon}\label{sec:4-1}

DiffPure \cite{nie2022diffusion} is a recent method that utilizes diffusion models to remove the imperceptible perturbations on the images by introducing Gaussian noise at a predetermined ratio \( t^* \) (ranging from 0 to 1) to adversarial images, followed by a denoising process via the reverse dynamics of diffusion models. Our study initially aimed to assess the applicability of DiffPure against patch attacks. The empirical investigations, illustrated in Fig. \ref{fig:new} demonstrate the inadequacy of DiffPure in countering patch attacks. 
This inefficacy stems from an inherent trade-off between purifying the adversarial perturbations (with a larger $t^*$) and preserving the image semantics (with a smaller $t^*$), making it impossible to find an appropriate noise ratio that can defend against adversarial patches.

In contrast, we find at a critical noise ratio of \( t^* \), a distinct pattern emerged: while the adversarial patches exhibited resistance to denoising, they also struggled to be restored, resulting in variable outcomes. Meanwhile, the remainder of the image retained its semantic integrity unscathed. This suggests that we can calculate the difference between various denoised images to identify the region of adversarial patches.
This observation, depicted through various examples in Fig. \ref{fig:observation}, leads to the \textbf{Adversarial Anomaly Perception (AAP)} phenomenon. The reason behind this phenomenon may be that adversarial patches are often intentionally crafted perturbations with a complexity far exceeding the noise present in real image datasets. Alternatively, it could be some meaningful sticker placed in an inappropriate location, signifying that the patch is out of context within the scenario. Diffusion models are trained to learn the probability distribution of real images, thus they struggle to fully adapt to the distribution of adversarial examples in its latent space, leading to the difference. 

The discovery of AAP provides insight into understanding how diffusion models can differentially respond to adversarial patches, which empowers the diffusion model to detect and localize adversarial patches through the analysis of distributional discrepancies, and further facilitates targeted restoration of specific patch-affected areas. It resolves the inherent trade-off between purging adversarial patches and preserving image authenticity. Leveraging AAP, we propose a unified diffusion-based defense framework \textbf{DIFFender}, employing a single diffusion model that both locates and restores patch attacks.

\subsection{DIFFender}\label{sec:4-2}

\noindent\textbf{Patch localization.}
DIFFender first performs accurate patch localization based on the above phenomenon of the diffusion model.
Given the adversarial image $\mathbf{x}_{a d v}$, we first add Gaussian noise to create a noisy image $\mathbf{x}_t$ with a certain noise ratio \( t^* \) (chosen as 0.5 in the experiments). Next, inspired by \cite{couairon2022diffedit}, we apply a text-guided diffusion model to obtain a denoised image $\mathbf{x}_p$ from $\mathbf{x}_t$ with a textual prompt $prompt_{L}$, and $\mathbf{x}_e$ with empty text.
% (representing $\mathbf{x}_{a d v}$ in latent form). 
We can estimate the mask region $\hat{\mathbf{M}}$ by taking the difference between the denoised images $\mathbf{x}_{p}$ and $\mathbf{x}_e$. 
However, the diffusion model incurs a significant time cost due to the time steps $T$ required. To address this issue, we directly predict the image $\mathbf{x}_0$ from the noisy image $\mathbf{x}_t$ with only one step, which saves $T$ times the processing time.

Although the one-step predicted results often exhibit discrepancies and increased blurriness compared to the original one, the differences between one-step predictions still align with the AAP phenomenon. In practice, we perform one-step denoising twice, obtaining two results: $\mathbf{x}_a$, the one guided by  $prompt_{L}$, and $\mathbf{x}_b$, the one guided by empty text to calculate the difference and binarize it, as:
\begin{equation}
\hat{\mathbf{M}} = \mathrm{Binarize}\left(\frac{1}{m} \sum\limits_{i=0}^m (\mathbf{x}_a^i -\mathbf{x}_b^i)\right),
\end{equation}
where we calculate the difference for $m$ times (set to 3 in the experiments)
to enhance stability and eliminate randomness. The $prompt_{L}$ can be hand-designed (e.g., "adversarial'') or automatically tuned as shown in~\cref{sec:4-3}.
% and reduce the computational complexity

\begin{figure}[!h]
  \centering
  % \fbox{\rule[-.5cm]{0cm}{4cm} \rule[-.5cm]{4cm}{0cm}}
  \includegraphics[width=0.55\linewidth]{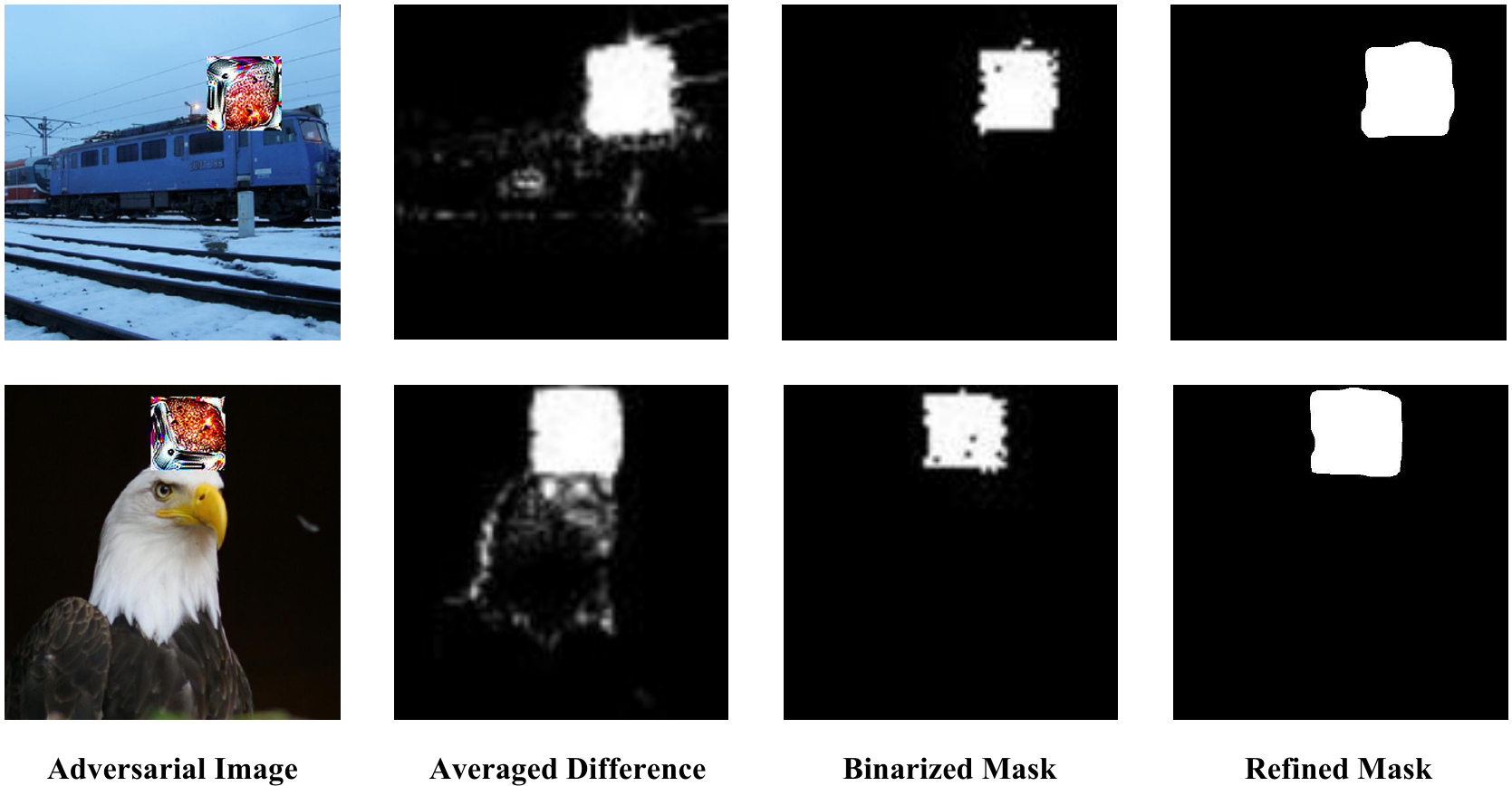} 

  \caption{\small To gain the final refined mask, the estimated differences are binarized, applied Gaussian smoothing and dilation operations.}
  \label{fig:refine}
\end{figure}

\noindent\textbf{Mask refining.}
As shown in Fig. \ref{fig:refine}, directly obtained averaged difference may sometimes result in minor inaccuracies. Therefore, we binarize the difference to gain the initial mask and then refine it by sequentially applying Gaussian smoothing and dilation operations, leading to a precise estimation of the patch region. The processed mask edges may slightly extend beyond the patch area, which helps maintain consistency in the patch restoration, thereby enhancing the overall performance of the defense pipeline.

\noindent\textbf{Patch restoration.}
After locating the patch region, DIFFender then restores it to eliminate the adversarial effects, while also considering preserving the overall coherence and quality of the image. In particular, we combine the estimated mask $\hat{\mathbf{M}}$ and $\mathbf{x}_{a d v}$ as inputs to the text-guided diffusion model with prompt $prompt_{R}$ to obtain a restored image $\mathbf{x}_r$. We follow the inpainting pipeline in Stable Diffusion \cite{Rombach_2022_CVPR} to process the mask, where a UNet is used with an additional five input channels to incorporate the estimated mask $\hat{\mathbf{M}}$. Similarly, $prompt_{R}$ can be manually set (e.g., "clean'') or automatically tuned. 

% To ensure the adversarial effect is removed in the images, the restoration performs the entire diffusion process, which takes several times the processing time of the localization. In practice, the restoration can be set to invoke only when attacks are detected, ensuring high efficiency (see Appendix~\ref{app:B.1} for details). 

\noindent\textbf{Unified defense model.}
 The aforementioned two stages have been meticulously integrated into one unified diffusion model (e.g., Stable Diffusion), driven by the critical AAP phenomenon. This intentional fusion allows us to harness the tight interplay between these stages, thereby enhancing the defense mechanism. As a direct consequence of our insights, we also introduced the prompt-tuning module, which encompasses the joint optimization of the entire pipeline.

\subsection{Prompt Tuning}\label{sec:4-3}

Following the aforementioned pipeline, leveraging vision-language pre-training, DIFFender is capable of efficiently performing zero-shot localization and restoration. While it is accurate in locating and restoring in most cases, subtle discrepancies may occur in certain challenging situations. Given that vision-language pre-training takes advantage of large-capacity text encoders to explore a vast semantic space~\cite{zhou2022learning}, to facilitate the effective adaptation of learned representations into adversarial defense tasks, we introduce the algorithm of prompt tuning.

\noindent\textbf{Learnable prompts.}
First, the textual vocabulary is replaced by learnable vectors. Thus, $prompt_{L}$ and $prompt_{R}$ can be transformed into vectors as follows:
\begin{equation}
\begin{split}
prompt_{L} =[V_L]_1 [V_L]_2 \ldots [V_L]_n; \\
prompt_{R} =[V_R]_1 [V_R]_2 \ldots [V_R]_n,
\end{split}
\end{equation}
where each $[V_L]_i$ or $[V_R]_i$ ($i \in \{1, \ldots, n\}$) is a vector of the same dimensionality as word embeddings. $n$ is a hyperparameter that specifies the number of context tokens, we set $n=16$ by default. The text content used to initialize $prompt_{L}$ and $prompt_{R}$ can be manually provided or randomly initialized.

\noindent\textbf{Tuning process.}
After obtaining the learnable vectors, we design three loss functions for prompt tuning, which jointly optimize the vectors to capture the characteristics of the adversarial regions and improve the defense performance. 

First, to accurately identify the adversarial regions, we employ cross-entropy loss by comparing estimated mask $\hat{\mathbf{M}}$ with ground-truth mask $\mathbf{M}$.
\begin{equation}
 L_{CE}(\mathbf{M},\hat{\mathbf{M}}) = -\sum_{i=1}^d\mathbf{M}_i\log(\hat{\mathbf{M}}_i),
\end{equation}
where $i$ indicates the $i$-th element of the mask.
Next, in the patch restoration module, our objective is to restore the mask region while eliminating the adversarial effect of the image. To ensure effective defense, we calculate the $\ell_1$ distance  between restored image $\mathbf{x}_r$ and clean image $\mathbf{x}$ as:
\begin{equation}  
L_1(\mathbf{x}_r, \mathbf{x})=\left| \mathbf{x}_r-\mathbf{x} \right|.
\end{equation}
Lastly, to verify that the adversarial effects have been eliminated, we draw inspiration from  \cite{liao2018defense} and \cite{zhang2018unreasonable} to make the high-level feature representations of the downstream classifiers between the restored image $\mathbf{x}_r$ and the clean image $\mathbf{x}$ close to each other. Specifically, we compute the $\ell_2$ distance between their feature representations weighted by a layer-wise hyperparameter as
% at each layer
\begin{equation}
d\left(\mathbf{x}_r, \mathbf{x}\right)=\sum_l \frac{1}{H_l W_l} \sum_{h, w}\left\|w_l \odot\left(\hat{y}_{r h w}^l-\hat{y}_{c h w}^l\right)\right\|_2^2,
\end{equation}
where $l$ denotes a certain layer in the network, $\hat{y}_r^l,\hat{y}_c^l \in \mathcal{R}^{H_l\times W_l \times C_l}$ are the unit-normalize results in the channel dimension, and vector $w^l\in\mathcal{R}^{C_l}$ is used for scaling activation channels. 

By summing three losses, the overall prompt tuning loss $L_{PT}$ is:
\begin{equation}
L_{PT} = L_{CE}(\mathbf{M},\hat{\mathbf{M}}) +L_1(\mathbf{x}_r, \mathbf{x})+d\left(\mathbf{x}_r, \mathbf{x}\right).
\end{equation}
We perform gradient descent to minimize $L_{PT}$ w.r.t. $prompt_L$ and $prompt_R$  for prompt tuning. The design of continuous representations also enables thorough exploration in the embedding space.

\noindent\textbf{Few-shot learning.}
During prompt tuning, we utilize a limited number of images for few-shot tuning. Specifically, DIFFender is tuned on a limited set of attacked images (8-shot in the experiments) from a single attack, but can learn optimal prompts that generalize well to other scenarios and attacks, which makes the tuning module effective and straightforward.

\section{Experiments}

\subsection{Experimental settings.}

\noindent\textbf{Datasets and Baselines.}
We consider ImageNet~\cite{deng2009imagenet} for evaluation against eight state-of-the-art defense methods: Image smoothing-based defenses, including JPEG~\cite{dziugaite2016study} and Spatial Smoothing~\cite{xu2017feature}, image completion-based defenses such as DW~\cite{hayes2018visible}, LGS~\cite{naseer2019local} and SAC~\cite{liu2022segment}, feature-level suppression defense FNC~\cite{yu2021defending}, alongside Jedi~\cite{tarchoun2023jedi}, a defense based on entropy. Additionally, we assess the diffusion purification defense, DiffPure~\cite{nie2022diffusion}. For classifiers, we consider two advanced classifiers: CNN-based Inception-v3~\cite{szegedy2016rethinking} and Transformer-based Swin-S~\cite{szegedy2016rethinking}.

\noindent\textbf{Adversarial attacks.}
We employ AdvP~\cite{brown2017adversarial} and LaVAN~\cite{karmon2018lavan}, which randomly select positions and optimize patches. GDPA~\cite{li2021generative}, which optimizes the patch's position and content to execute attacks, and natural-looking attack RHDE~\cite{wei2022adversarial}, which utilizes realistic stickers and searches for their optimal positions to launch attacks. To implement adaptive attacks, we approximate gradients using the BPDA~\cite{athalye2018obfuscated} to conduct BPDA+AdvP and BPDA+LaVAN attacks, which implies that the defense methods are white-box against the attacks. The number of iterations for the attacks is set to 100 with patch size 5\% of the input image. For adapting the attack on DIFFender, we use an additional Straight-Through Estimator (STE)~\cite{yin2019understanding} during backpropagation through thresholding operations. 
% Additionally, due to the randomness introduced by the diffusion processes, we employ the Expectation over Transformation (EOT)~\cite{athalye2018obfuscated}.

\noindent\textbf{Evaluation metrics.}
We evaluate the performance of defenses under standard accuracy and robust accuracy. Due to the computational cost of adaptive attacks, unless otherwise specified, we assess the robust accuracy on a subset of 512 sampled images from the test set. To facilitate the observation, we ensure that the selected subset consists of images correctly classified. 
 % for details

\subsection{Evaluation on ImageNet}

\begin{table*}[t]
\centering
\small
\caption{\small Accuracy (\%) against attacks on ImageNet by Inception-v3 and Swin-S.}

  \scalebox{0.88}{
  \begin{tabular}{c|c|cc|cc|c|cc|cc}
  \toprule
               Models & \multicolumn{5}{c|}{Inception-v3}   & \multicolumn{5}{c}{Swin-S}         \\

\cmidrule(lr){1-11} 
% \cmidrule(lr){7-11}
&  & \multicolumn{2}{c|}{Adaptive} & \multicolumn{2}{c|}{Non-adaptive} &  & \multicolumn{2}{c|}{Adaptive} & \multicolumn{2}{c}{Non-adaptive} \\
\diagbox[height=0.6cm]{Defense }{Attack}       & Clean & AdvP & LaVAN & GDPA & RHDE & Clean & AdvP & LaVAN & GDPA & RHDE \\ \midrule
Undefended     & 100.0 & 0.0 & 8.2 & 64.8 & 39.8 & 100.0 & 1.6 & 3.5 & 78.1 & 51.6 
\\
JPEG~\cite{dziugaite2016study}              & 48.8 & 0.4 & 15.2 & 64.8 & 13.3 & 85.2 & 0.8 & 5.9 & 77.0 & 38.7 
\\
SS~\cite{xu2017feature}              & 72.7 & 1.2 & 14.8 & 57.8 & 16.4 & 86.3 & 2.3 & 5.5 & 68.8 & 34.8 
\\
DW~\cite{hayes2018visible}             & 87.1 & 1.2 & 9.4 & 62.5 & 28.5 & 88.3 & 0.0 & 5.1 & 77.3 & 66.0 
\\
LGS~\cite{naseer2019local}            & 87.9 & 55.5 & 50.4 & 67.2 & 49.6 & 89.8 & 65.6 & 59.8 & 82.0 & 69.1 
\\
FNC~\cite{yu2021defending}            & 91.0 & 61.3 & 64.8 & 66.4 & 46.5 & 91.8 & 6.3 & 7.4 & 77.0 & 63.7 
\\
DiffPure~\cite{nie2022diffusion}         & 65.2 & 10.5& 15.2& 67.6& 44.9& 74.6& 18.4& 26.2& 77.7& 62.3
\\
SAC~\cite{liu2022segment}           & \textbf{92.8} &84.2   &65.2   &68.0   &41.0   &93.6   &92.8   &84.6   &79.3   &54.9 
\\
Jedi~\cite{tarchoun2023jedi}           & 92.2 & 67.6 & 20.3 & 74.6 & 47.7 & 93.4 & 89.1 & 12.1 & 78.1 & 67.6 
\\

DIFFender & 91.4 & \textbf{88.3} & \textbf{71.9} & \textbf{75.0} & \textbf{53.5} & \textbf{93.8} & \textbf{94.5} & \textbf{85.9} & \textbf{82.4} & \textbf{70.3} \\

\bottomrule
\end{tabular}
 }
  \label{tab:1}
\end{table*}

\begin{figure*}[h]
  \centering
  % \fbox{\rule[-.5cm]{0cm}{4cm} \rule[-.5cm]{4cm}{0cm}}
  \includegraphics[width=0.95\linewidth]{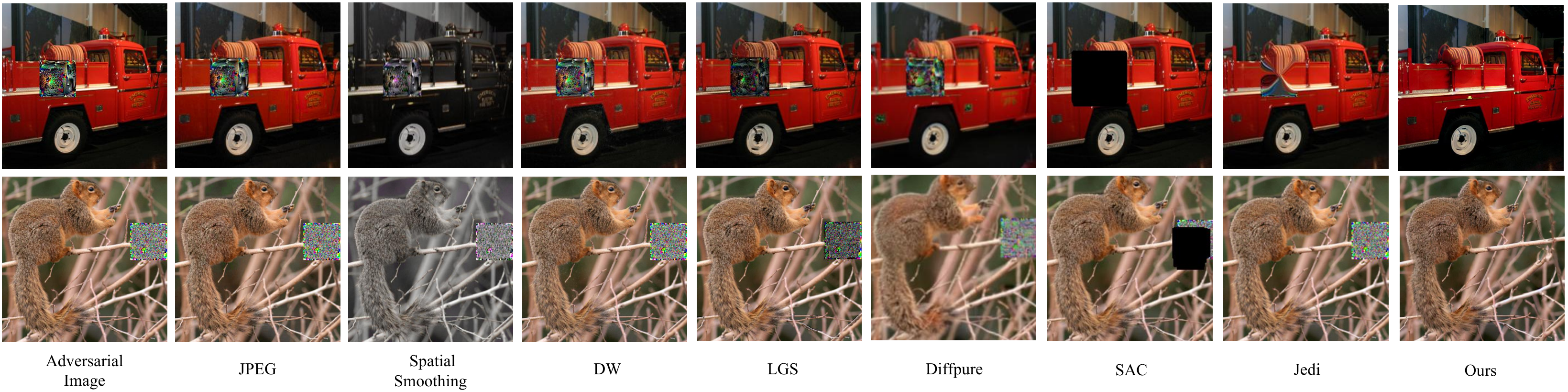}
    % \vspace{-0.6cm}
  \caption{\small Visualization with examples from ImageNet. The restored images of DIFFender exhibit no residual signs of the adversarial patch, and the restored details are remarkable (e.g., the recovery of tree branches in the second column of images).}
  \label{fig:3}
  % \vspace{-0.6cm}
\end{figure*}

\noindent\textbf{Experimental results.}
Tab.~\ref{tab:1} presents the experimental results, where the highest accuracy is highlighted in bold. Based on these results, we draw the following conclusions:

(1) DIFFender outperforms in defense effectiveness. Under adaptive attacks utilizing gradients, such as the BPDA+AdvP and BPDA+LaVAN, DIFFender exhibits exceptional performance, even only involving an 8-shot process. Other attacks like GDPA may not achieve strong attack effectiveness, but DIFFender still attains the highest robust accuracy. This can be attributed to that DIFFender is built upon the unified diffusion framework. Empowered by the AAP phenomenon, the diffusion model can effectively locate and remove adversarial areas while ensuring a high-quality and diverse restoration that closely follows the underlying distribution of clean data. Additionally, the inherent stochasticity in the diffusion model allows for robust stochastic defense mechanisms~\cite{he2019parametric}, which makes it a well-suited "defender" for adaptive attacks.

(2) Image processing defense methods, such as JPEG, SS, and DW, experience a significant decrease in robust accuracy under adaptive attacks. This can be attributed to the algorithms' gradients can be easily obtained. Other methods, such as LGS, FNC, SAC and Jedi, consider the robustness against adaptive attacks. For instance, FNC achieves respectable robust accuracy on Inception-v3. However, its defense effectiveness diminishes when applied to the Swin-S. This is because the feature norm clipping layer proposed is specifically designed for handling CNN feature maps, while DIFFender exhibits excellent generalization capabilities that can extend to different classifiers.

(3) DIFFender demonstrates generalizability to unseen attacks. In the experiments, DIFFender only undergoes 8-shot prompt tuning specifically for the AdvP method yet achieves promising results under other attacks as well. For Jedi, it has strong robustness against several attack methods, such as AdvP, but its robust accuracy has significantly decreased under other methods, like LaVAN. This might be because the autoencoder used by Jedi is trained under a specific style and cannot generalize well.

% However, it still maintains good performance when facing other attack methods. 

(4) For the naturalistic attack RHDE, it poses a lesser threat to classifiers compared to adaptive meaningless attacks. However, it introduces a more significant challenge to defense methods, likely due to its utilization of irregular, more naturally-appearing patches. Nevertheless, DIFFender still achieves the best defense results without prior exposure to RHDE patches. Moreover, DIFFender is adaptable; with the prompt tuning module, a few-shot tuning can be employed to enhance performance specifically against naturalistic patch attacks.

(5) When defending against global perturbations with $\ell_p$-norm constraints, DiffPure achieves excellent results. However, it performs poorly when facing patch attacks. Specifically, in Tab.~\ref{tab:1}, when tested against AdvP and LaVAN, the Inception-v3 model purified by DiffPure maintains robust accuracy rates of 10.5\% and 15.2\%, respectively. This aligns with our observations in~\cref{sec:4-1}.

\noindent\textbf{Visualization.}
Fig.~\ref{fig:3} presents the defense results of the defense methods against patch attacks. Since FNC suppresses the feature maps during the inference stage, it is not shown in Fig.~\ref{fig:3}. Other methods, such as JPEG and DW, only exhibit minor changes in the reconstructed images and fail to defend against adaptive attacks. After Spatial Smoothing defense, the images show color distortion and are still vulnerable to attacks. In the case of the LGS method, the patch area is visibly suppressed, which improves the robust accuracy to some extent, but the patch is not completely eliminated. For Jedi and SAC, their localization algorithm fails under certain scenarios, as the second line in Fig.~\ref{fig:3}. And the restored results of Jedi cannot achieve complete recovery. On the other hand, the restored images of DIFFender no longer show any traces of the patch, and the restored details are remarkable.

\subsection{ Ablation studies and additional results}\label{sec:5-2}

\noindent\textbf{Impact of loss functions.}
To evaluate the effectiveness of different losses, we conduct separate tuning experiments by removing loss functions $L_{CE}$, $L_1$, and $d$ separately, as presented in Tab.~\ref{tab:loss}, where we observe that the robust accuracy significantly decreases when optimizing only the Restoration module without optimizing $L_{CE}$ due to the performance loss in the localization, although it led to an improvement in clean accuracy. On the other hand, removing $L_1$ results in a noticeable decrease in clean accuracy, as images cannot be well restored. Eliminating either the \(d\) or $L_1$ causes a slight drop in robust accuracy. Finally, DIFFender, which incorporates all three loss functions, achieves the highest robust accuracy, demonstrating the importance of joint optimization and close interaction between the two modules for the overall performance of DIFFender.

\noindent\textbf{Impact of patch size.}
We conducted experiments under patch attacks of varying sizes, using patches generated by AdvP ranging from 0.5\% to 15\% in size, and compared them with the state-of-the-art SAC and Jedi. As shown in Tab.~\ref{Patch}, DIFFender exhibits better generalization to patches of various sizes, benefitting from vision-language pre-training, whereas Jedi and SAC are more sensitive to patch size. Notably, DIFFender was only prompt-tuned for patches of 5.0\% size.

\begin{table}[!t]
\centering%
\begin{minipage}[b]{0.48\textwidth}
  \centering

\renewcommand\arraystretch{0.98}
\caption{\small Ablation study for different loss functions of DIFFender.}
  \scalebox{0.75}{
        \begin{tabular}{ccc|ccccc} 
            \toprule
                   \multicolumn{1}{c}{} & \multicolumn{1}{c}{} & \multicolumn{1}{c}{} & \multicolumn{5}{c}{Inception-v3}        \\ \hline
                $L_{CE}$ & $L_1$& $d$& Clean & AdvP & LaVAN & GDPA & RHDE \\ \hline
                 & \checkmark & \checkmark & \textbf{91.8} & 76.2 & 66.0 & 72.3 &49.2 
\\ 
                \checkmark & ~ & \checkmark & 88.3 & 87.1 & 69.5 & 73.8 &52.7 
\\ 
                \checkmark & \checkmark & ~ & 90.2 & 87.1 & 69.1 & 73.0 &52.0 
\\ 
                \checkmark & \checkmark & \checkmark  & 91.4 & \textbf{88.3} & \textbf{71.9} &  \textbf{75.0} &\textbf{53.5} \\ 
            \bottomrule
          \end{tabular}
         }
          \label{tab:loss}
\end{minipage}%
\hspace{3mm}%
\begin{minipage}[b]{0.48\textwidth}
\centering
\caption{\small Accuracy against attacks of varying patch sizes by Inception-v3.}
\scalebox{0.80}{
\begin{tabular}{c|ccccc}
\toprule
Size & 0.5\% & 1.0\%   & 5.0\%   & 10.0\%  & 15.0\%  \\ \midrule
Undefended & 64.3 & 50.8   & 0.0   & 0.0  & 0.0    \\
SAC & 81.8 & 83.8   & 84.2   & 60.9  & 34.8   \\
Jedi & 61.7 & 56.4   & 67.6   & 42.2  & 33.8   \\
DIFFender & \textbf{86.1} & \textbf{87.3}   & \textbf{88.3}   & \textbf{70.5}  & \textbf{56.6}   \\ \bottomrule
\end{tabular}
}
  \label{Patch}
\end{minipage}
\end{table}

\begin{table*}[!t]
% \centering
\small
 \centering
\caption{\small Ablation study for restoration modules in DIFFender. "NR" denotes "No Restoration Process". }

  \scalebox{0.86}{
  \begin{tabular}{c|ccccc|ccccc}
  \toprule
               & \multicolumn{5}{c}{Inception-v3}   & \multicolumn{5}{c}{Swin-S}         \\
\textbf{Defense}        & Clean & AdvP & LaVAN & GDPA & RHDE & Clean & AdvP & LaVAN & GDPA & RHDE \\ \midrule
DIFFender (NR) & 86.3 & 84.0 & 66.8 & 69.5 & 48.0 & 88.7 & 92.2 & 81.8 & 78.9 & 69.1 \\
DIFFender & \textbf{91.4} & \textbf{88.3} &\textbf{71.9} & \textbf{75.0} & \textbf{53.5}& \textbf{93.8}& \textbf{94.5} & \textbf{85.9} & \textbf{82.4} & \textbf{70.3}  \\

\bottomrule
\end{tabular}
}
  \label{tab:Restoration}
\end{table*}

\begin{table*}[!t]
 \centering
\small
\caption{\small Ablation study for different prompt forms. "EP" and "MP" represent "Empty Prompt" and "Manual Prompt".}
  \scalebox{0.86}{
  \begin{tabular}{c|ccccc|ccccc}
  \toprule
               & \multicolumn{5}{c}{Inception-v3}   & \multicolumn{5}{c}{Swin-S}         \\
\textbf{Defense}        & Clean & AdvP & LaVAN & GDPA & RHDE & Clean & AdvP & LaVAN & GDPA & RHDE \\ \midrule
DIFFender (EP)& 89.1  & 76.4 & 66.8  & 71.1 & 47.0 & 93.2  & 89.8 & 81.4  & 79.3 & 65.7 \\
DIFFender (MP) & 87.3  & 77.9 & 68.2  & 70.3 & 47.8& 92.2  & 91.2 & 82.4  & 77.0 & 67.6 \\
DIFFender & \textbf{91.4} & \textbf{88.3} &\textbf{71.9} & \textbf{75.0} & \textbf{53.5} & \textbf{93.8} & \textbf{94.5} & \textbf{85.9} & \textbf{82.4} & \textbf{70.3} \\
\bottomrule
\end{tabular}
}
  \label{tab:6}
\end{table*}

\begin{table*} [!t]
\small
  \caption{\small Transferability of DIFFender on ResNet50 and ViT-B-16 for ImageNet.}
  \centering
    \scalebox{0.92}{
  \begin{tabular}{c|ccccc|ccccc}
    \toprule

                 & \multicolumn{4}{c}{ResNet-50}      & \multicolumn{4}{c}{ViT-B-16}       \\ 
\textbf{Defense} & Clean         & AdvP    & LaVAN       & GDPA & RHDE & Clean         & AdvP     & LaVAN        & GDPA & RHDE  \\ \hline
Undefended      & 100.0 & 0.0  & 14.8  & 73.8  &37.1 & 100.0 & 1.2  & 2.0   & 76.2 &52.0 \\
Jedi            & 82.8 & 46.9 & 8.6 & 70.7 & 45.5 & 89.8 & 83.8 & 14.8 & 76.8 & 59.8 \\
DIFFender         & \textbf{83.6}  & \textbf{83.2} & \textbf{55.9}  & \textbf{76.2}  &\textbf{53.5} & \textbf{91.0}  & \textbf{88.3} & \textbf{85.2}  & \textbf{78.9} &\textbf{68.0} \\
    \bottomrule
  \end{tabular}
  }
  \label{tab:2}

\end{table*}

\noindent\textbf{Impact of restoration module.}
To verify the necessity of restoration, we remove the patch restoration and set the value in the $\hat{\mathbf{M}}$ region to zero. Experimental results in Tab.~\ref{tab:Restoration} show that the inclusion of patch restoration ensures better DIFFender performance. This is because patches may occasionally obscure crucial areas of an image, resulting in a loss of semantic information. The restoration step can address this issue by recovering lost semantics, aiding classifiers in overcoming challenging scenarios. Furthermore, longer diffusion steps introduce more randomness, which preserves accuracy against adaptive attacks. Consequently, we conclude that the patch restoration is indeed necessary.

\noindent\textbf{Impact of Prompt Tuning.}
In Tab.~\ref{tab:6}, DIFFender with prompt-tuning is compared with the "Empty prompt" and "Manual prompt" versions of DIFFender. For DIFFender with manual prompts, we set $prompt_L$ = "adversarial" and $prompt_R$ = "clean". The prompt-tuned DIFFender shows a notable improvement in robust accuracy compared to the other two zero-shot versions, despite exposure to only a few attacked images, underscoring the effectiveness of prompt-tuning.

\noindent\textbf{Cross-model transferability.}
We apply only 8-shot prompt-tuning exclusively on Inception-v3. Subsequently, the transferability of DIFFender is tested on diverse classifiers, including the CNN-based ResNet50~\cite{he2016deep} and Transformer-based ViT-B-16~\cite{dosovitskiy2020image}. The results are detailed in Tab.~\ref{tab:2}, underscore DIFFender's ability to maintain robust accuracy across novel classifiers, demonstrating its potent generalization capacity.

\begin{minipage}[c]{0.48\textwidth}
\centering%
\includegraphics[width=0.72\linewidth]{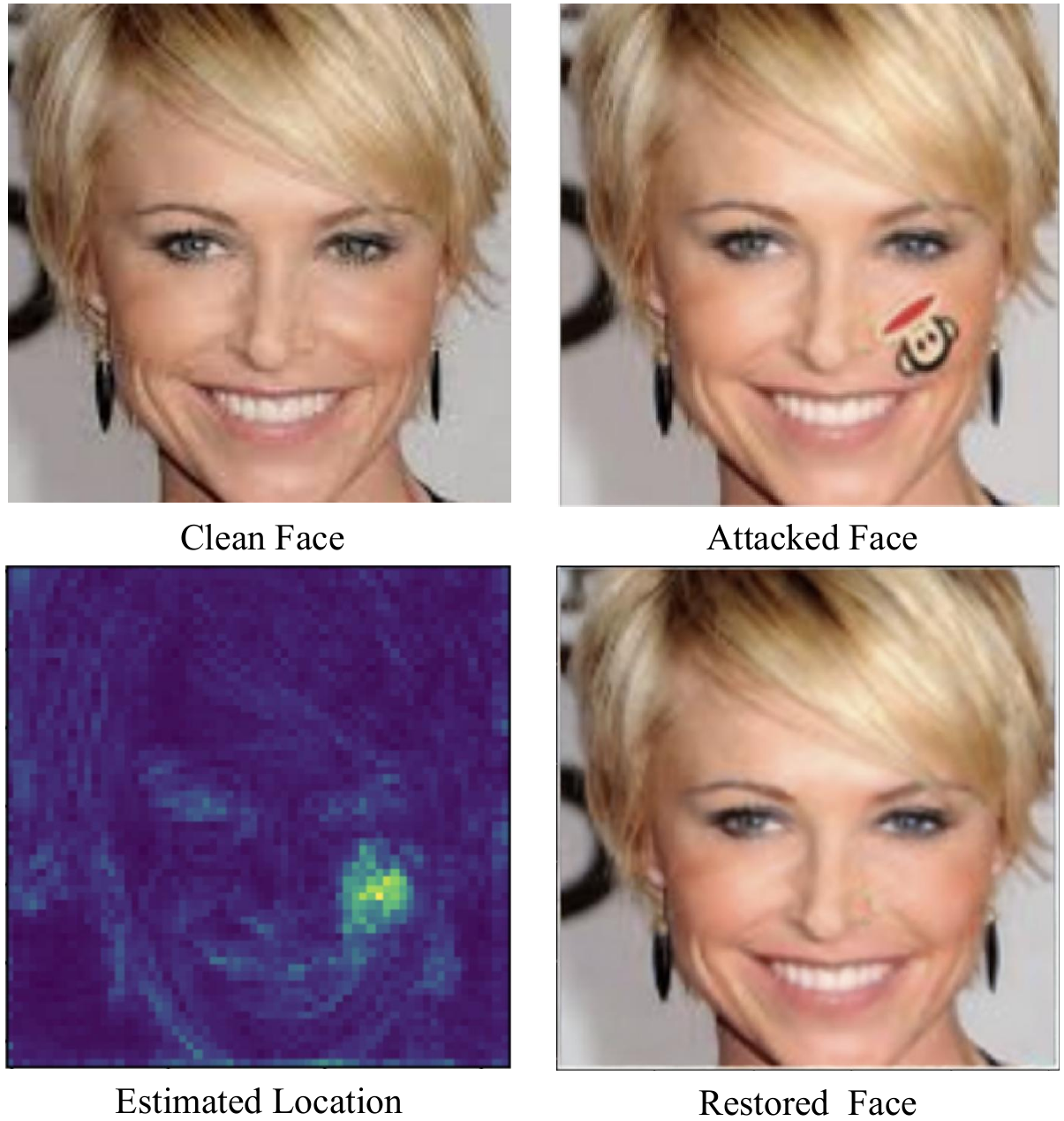}
\captionof{figure}{\small Visualization with examples from LFW attacked naturally by RHDE, localized and restored by DIFFender.}
\label{fig:face}
\end{minipage}%
% \hspace{2mm}%
\begin{minipage}[c]{0.48\textwidth}
  \captionof{table}{\small Accuracy against patch attacks on LFW by FaceNet.}
    \centering
      \scalebox{0.82}{
    \centering
    \begin{tabular}{c|ccc}
    \toprule

                 & \multicolumn{3}{c}{FaceNet} \\ 
\textbf{Defense} & Clean    & GDPA    & RHDE  \\ \midrule
Undefended      & 100.0 & 56.3 & 42.8 
\\
JPEG~\cite{dziugaite2016study}             & 44.1 & 16.8 & 17.8 \\
SS~\cite{xu2017feature}                  & 19.9 & 8.2 & 3.5 \\
DW~\cite{hayes2018visible}               & 37.1 & 15.2 & 7.2 \\
LGS~\cite{naseer2019local}               & 60.9 & 71.9 & 53.5 \\
FNC~\cite{yu2021defending}                & 100.0 & 39.8 & 39.3 \\
SAC~\cite{liu2022segment}                 & 100.0 & 77.3 & 43.2 \\
Jedi~\cite{tarchoun2023jedi}               & 100.0 & 74.2 & 43.9 \\
DIFFender (EP)        & 100.0 & 79.3 & 57.2 \\
DIFFender (MP)         & 100.0 & 77.0 & 57.2 \\
DIFFender         & \textbf{100.0} & \textbf{81.1} & \textbf{60.7}  \\
    \bottomrule
  \end{tabular}}
  \label{tab:7}
\end{minipage}\\

\subsection{Extension in Face Recognition.}\label{sec:5-3}
\noindent\textbf{Experimental settings.}
Facial expressions in human faces introduce a rich diversity, together with external factors such as lighting conditions and viewing angles, making face recognition a challenging task. We conducted experiments on the LFW~\cite{huang2008labeled}, and employed two patch attacks: RHDE~\cite{wei2022adversarial} and GDPA~\cite{li2021generative}.

\noindent\textbf{Experimental results.}
The results on the LFW dataset are presented in Tab.~\ref{tab:7}. DIFFender achieves the highest robust accuracy under both the GDPA and RHDE while ensuring clean accuracy. It is important to note that DIFFender is not tuned specifically for facial recognition. This further demonstrates the generalizability of DIFFender across different scenarios and attack methods. In contrast, JPEG, SS, and the FNC method obtained low robust accuracies. This is because in the specific context of facial recognition, the classifier focuses more on crucial local features, and preprocessing the entire image can disrupt these important features. Fig.~\ref{fig:face} illustrates the results of DIFFender against face attacks. It can be observed that DIFFender accurately identifies the location of the natural patch and achieves excellent restoration.

\subsection{Extension in Physical World.}

\begin{figure*}[t]
  % \fbox{\rule[-.5cm]{0cm}{4cm} \rule[-.5cm]{4cm}{0cm}}
  \includegraphics[width=0.75\linewidth]{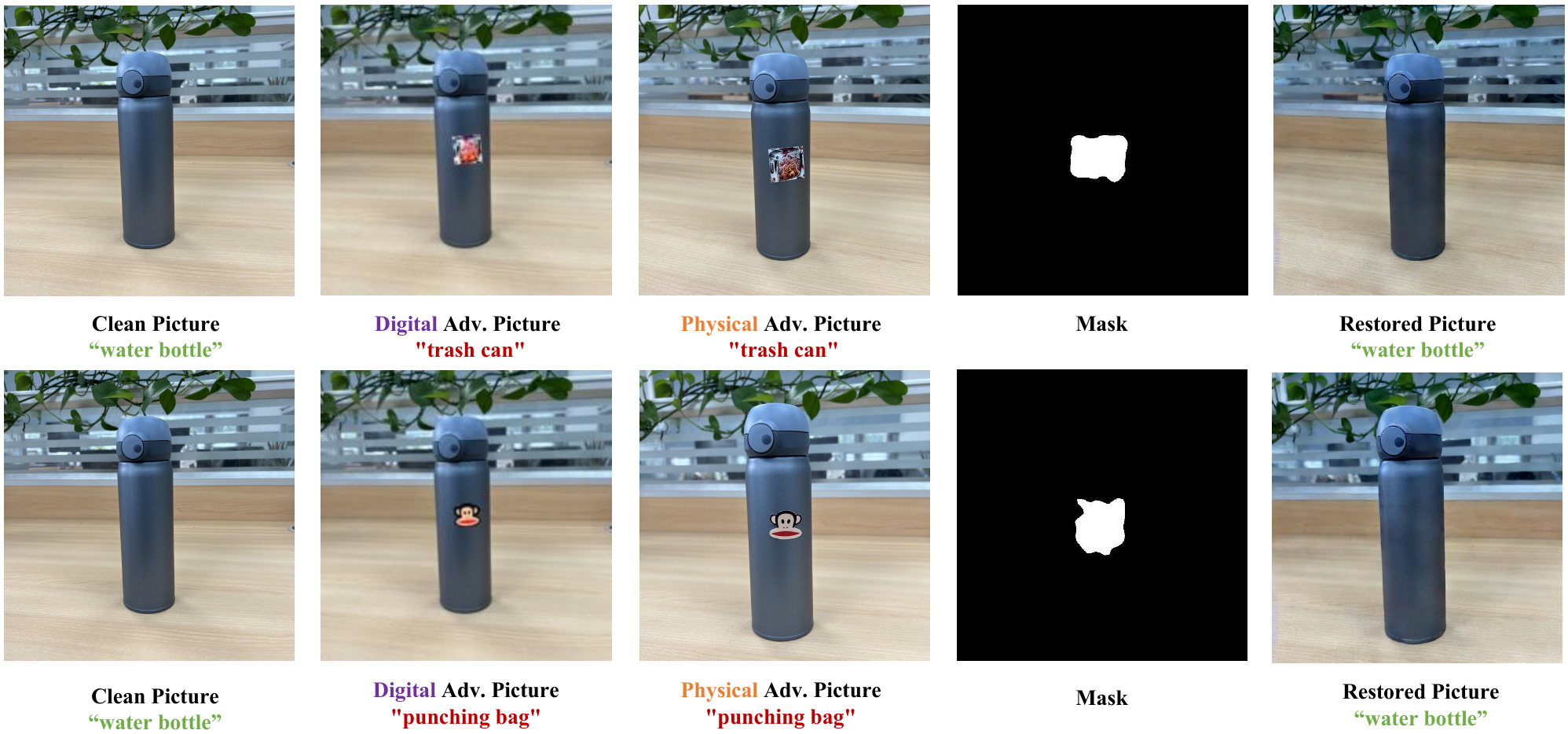}
  \centering
  \caption{\small Physical world defense demonstrations of DIFFender against meaningless and natural patch attacks. The mask edges may extend slightly beyond the patch region, aid in restoring the patch, and help maintain consistency in the restored image.}
  \label{fig:6}
\end{figure*}

We additionally conduct further experiments in the physical world, where we select 10 common object categories from ImageNet and perform two types of patch attacks (natural and meaningless)~\cite{wei2022physically}. Our approach involves generating digital-world attack results first, then placing stickers on real-world objects in the same positions.  We test DIFFender under various conditions, including different angles (rotations) and distances. Qualitative results are depicted in Fig.~\ref{fig:6}, while quantitative results are presented in Tab.~\ref{tab:8}, where each configuration is based on 256 frames successfully classified images from the 10 objects selected. Based on the results, we see that DIFFender demonstrates substantial defensive capabilities across various physical alterations, maintaining its efficacy in real-world scenarios.

\begin{table}[t]
\caption{\small Quantitative result of meaningless physical attacks on the Inception-v3 at different angles and distances. }
\centering
  \label{tab:8}
\centerline{
\scalebox{0.86}{
\begin{tabular}{c|ccccc}
 \toprule
               & 0°   & yaw ±15° & yaw ±30° & pitch ±15° & distance \\ \midrule
Undefended     & 28.9& 34.8& 41.8& 36.7& 35.9\\
Jedi           & 61.7 & 57.8 & 66.4 & 63.3 & 62.1 \\
DIFFender & \textbf{80.9} & \textbf{76.6} & \textbf{77.7} & \textbf{75.4} & \textbf{73.8} \\ \bottomrule
\end{tabular} 
}
}
\end{table}

\section{Discussion and Conclusion}

We propose \textbf{DIFFender}, a novel defense framework harnessing a pre-trained unified diffusion model for dual roles in the localization and restoration of patch attacks, empowered by the discovery of the Adversarial Anomaly Perception (AAP) phenomenon. Additionally, we design a few-shot prompt-tuning algorithm to facilitate simple and efficient tuning, thus eliminating the need for extensive retraining. To validate the robust performance of DIFFender, we conduct experiments on image classification, face recognition, and further the physical world scenarios. Our findings demonstrate that DIFFender exhibits exceptional robustness even under adaptive attacks and extends the generalization capability of pre-trained large models to various scenarios, diverse classifiers, and multiple attack methods, requiring only a few-shot prompt-tuning. We prove that DIFFender effectively reduces the success rate of patch attacks while producing realistic restored images, promising a wide spectrum of diffusion model applications and inspiring future explorations in the domain.

There are several avenues for further exploration. While we have designed acceleration techniques to expedite diffusion-based methods, further acceleration can be achieved by adopting advanced acceleration sampling methods. Another potential direction for exploration is to extend the DIFFender framework to other modalities of adversarial attack defenses, such as video data.

\section*{Acknowledgement}
This work was supported by the Project of the National Natural Science Foundation of China (No. 62076018, 92370124, 62276149, 92248303), the Fundamental Research Funds for the Central Universities, and Tsinghua-Alibaba Joint Research Program. Y. Dong
was also supported by the China National Postdoctoral Program for Innovative Talents and CCF-BaiChuan-Ebtech Foundation Model Fund.

\bibliographystyle{splncs04}
\bibliography{main}

\clearpage
\setcounter{page}{1}
% \maketitlesupplementary

% ---------------------------------------------------------------

\end{document}